\documentclass[pdflatex,sn-mathphys-num]{sn-jnl}

\usepackage{hyperref}
\usepackage{bookmark}
\usepackage{subfigure}
\usepackage{times}
\usepackage{graphicx}%
\usepackage{multirow}%
\usepackage{amsmath,amssymb,amsfonts}%
\usepackage{amsthm}%
\usepackage{mathrsfs}%
\usepackage[title]{appendix}%
\usepackage{xcolor}%
\usepackage{textcomp}%
\usepackage{manyfoot}%
\usepackage{booktabs}%
\usepackage{algorithm}%
\usepackage{algorithmicx}%
\usepackage{algpseudocode}%
\usepackage{listings}%

\newtheorem{definition}{Definition}%

\begin{document}

\title[Article Title]{Attentive Graph Enhanced Region Representation Learning}


\author[1]{\fnm{Weiliang} \sur{Chen}}\email{chanweiliang@s.hlju.edu.cn}

\author*[1]{\fnm{Qianqian} \sur{Ren}}\email{renqianqian@hlju.edu.cn}

\author[2]{\fnm{Jinbao} \sur{Li}}\email{lijinb@sdas.org}

\affil*[1]{\orgdiv{Department of Computer Science and Technology}, \orgname{Heilongjiang University}, \orgaddress{\street{74 Xuefu Rd}, \city{Harbin}, \postcode{150080}, \state{Heilongjiang}, \country{China}}}

\affil[2]{\orgdiv{Shandong Artificial Intelligence Institute}, \orgname{Qilu University of Technology}, \orgaddress{\street{19 Keyuan Rd}, \city{Jinan}, \postcode{250013}, \state{Shandong}, \country{China}}}


\abstract{Representing urban regions accurately and comprehensively is essential for various urban planning and analysis tasks. Recently, with the expansion of the city, modeling long-range spatial dependencies with multiple data sources plays an important role in urban region representation. In this paper, we propose the Attentive Graph Enhanced Region Representation Learning (ATGRL) model, which aims to capture comprehensive dependencies from multiple graphs and learn rich semantic representations of urban regions. Specifically, we propose a graph-enhanced learning module to construct regional graphs by incorporating mobility flow patterns, point of interests (POIs) functions, and check-in semantics with noise filtering. Then, we present a multi-graph aggregation module to capture both local and global spatial dependencies between regions by integrating information from multiple graphs. In addition, we design a dual-stage fusion module to facilitate information sharing between different views and efficiently fuse multi-view representations for urban region embedding using an improved linear attention mechanism. Finally, extensive experiments on real-world datasets for three downstream tasks demonstrate the superior performance of our model compared to state-of-the-art methods. }




\maketitle

\section{Introduction}\label{sec1}
In recent years, urban region representation has gained significant importance across various domains, leveraging its ability to provide valuable and essential information. This information has been utilized in several applications, such as socio-demographic feature prediction \cite{Jean_2018_Tile2Vec, Wang_2020_Urban2Vec}, crime prediction \cite{Wang_2017_HDGE, Zhang_2020_MVGRE, Wu_2022_MGFN}, economic growth prediction\cite{Hui_2020_Predicting} and land usage classification \cite{Yao_2018_ZeMob, region2vec}.
Urban region representation learning refers to the process of capturing and encoding the essential characteristics and features of urban regions in a structured and meaningful manner. It involves the extraction and transformation of raw urban data into a representation that encapsulates the spatial, temporal, and semantic information of a given region.
With the growing availability and accessibility of urban data, there has been a growing interest in developing effective techniques for urban region representation learning. However, it could be largely affected by the complicated spatial and temporal factors, and data heterogeneity and thus still suffers from many challenges.

One of the key points in urban region representation that must be solved is to model complicated correlations with heterogeneous data. Indeed, various urban representation models have been proposed, demonstrating their effectiveness in enhancing the scope and accuracy of region representation learning\cite{Cui_2017_Survey, Abu-El-Haija_2018_Watch, Kipf_2016_Variational, Gilmer_2017_Neural}. Among these models, graph neural networks (GNNs) have garnered attention for their ability to learn low-dimensional embeddings of graph-structured data, including urban regions. GNNs have become a popular choice for capturing complex dependencies among multiple sources of urban data, leveraging the inherent graph connections\cite{Zhang_2017_RPA, Wang_2017_HDGE, Zhang_2020_MVGRE, Chang_2020_Understanding, Wu_2022_MGFN, Luo_2022_ProfilingFramework, Zhang_2023_AutoST}.
For instance, Wang et.al \cite{Wang_2017_HDGE} proposes to construct graph structures where regions are represented by nodes in the graph.
Zhang et.al \cite{Zhang_2017_RPA, Yao_2018_ZeMob} incorporates various factors such as urban regions, time, human movement activities, and others to create a heterogeneous graph that effectively captures regional embeddings.

\begin{figure}
\centering\includegraphics[scale=.23]{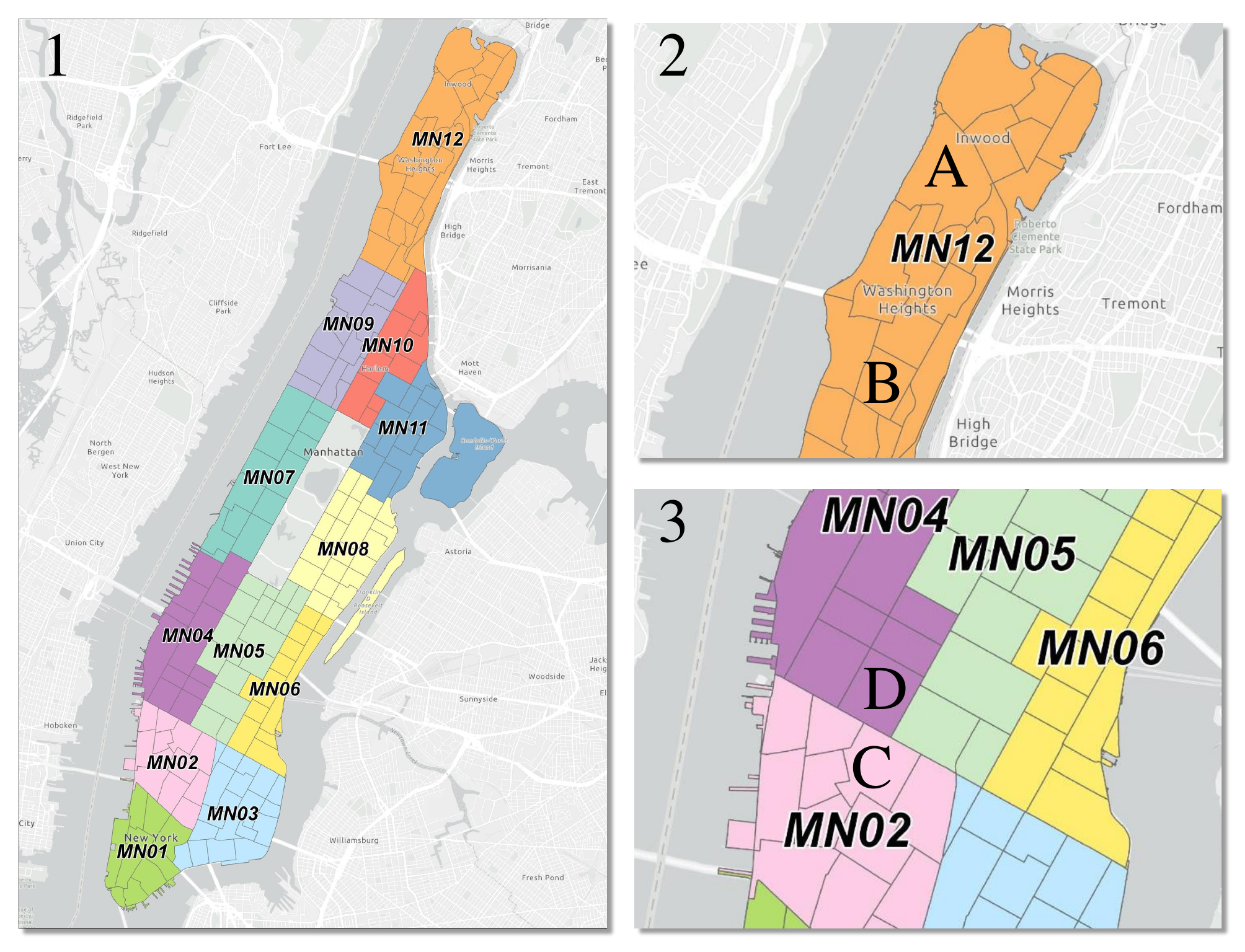}
\caption{The Borough of Manhattan in New York City serves as the study area for this paper. As shown in sub-figure 1, the Community Boards \cite{Berg__New} have divided Manhattan into 12 districts based on land use. To conduct a more comprehensive analysis, the study area is divided into 180 regions.}
\label{longrange}
\end{figure}

Recently, many studies have adopted a multi-view perspective to learn graph embeddings, incorporating diverse sources of information \cite{Zhang_2020_MVGRE, Wu_2022_MGFN, Luo_2022_ProfilingFramework}.
To effectively capture the important correlations among various regions and fuse multiple view features, the graph attention network (GAT) \cite{Velickovic_2018_GAT} and self-attention mechanism \cite{Vaswani_2017_Attention} are often used \cite{Zhang_2020_MVGRE, Luo_2022_ProfilingFramework}. However, they have limits of complexity and ignore
potential dependencies.
Although existing GNNs-based solutions have obtained promising results for urban region embedding,  the following three challenges have not been well addressed.
\begin{enumerate}
\item \textbf{Data Heterogeneity.} Urban data comes from various sources, such as geospatial data, transportation data, social media data, and demographic data. These data sources often have different formats, resolutions, and quality levels. Integrating and harmonizing these heterogeneous data types to create a unified representation poses a significant challenge.
\item \textbf{Noise and incompleteness of urban data.} 
In real-world applications, environmental factors like sensor failures and measurement errors introduce noise and incompleteness to urban data. This leads to the inclusion of noisy information in the constructed graph structure from the original data. Consequently, capturing dependencies within these noisy graphs can significantly impact the quality of the representation.
\item \textbf{Long-range dependencies among regions.}
Many existing models focus on local dependencies among neighboring regions, potentially overlooking the significance of capturing relationships that extend over longer ranges. For instance, as illustrated in Figure.~\ref{longrange}, Regions A and B are situated in the same district(orange area), indicating a functional partition of the city. Despite not being immediate neighbors, their spatial proximity within the district suggests a strong relationship that should be captured in the representation.  On the other hand, Regions C and D, while neighboring regions, belong to different districts, indicating that the district boundary may introduce a distinct set of relationships and dependencies.
\end{enumerate}

To address these challenges, we present ATGRL, a novel urban region profiling model that employs a heterogeneous graph neural framework accommodating diverse data sources such as human mobility, POI functions, and check-in semantics. Our approach begins with the graph-enhanced learning module for constructing graphs from various features while also incorporating lightweight noise filtering techniques. Then, the multi-graph aggregation module is designed to capture complex dependencies and interactions among different features and regions. Subsequently, we introduce a dual-stage graph fusion module to combine information from different graphs and facilitate information sharing across views.  
Our work makes the following notable contributions:

\begin{itemize}
\item We propose the idea of graph-enhanced learning, which recognizes the heterogeneity of urban data and proposes a framework that models and learns individual graphs for each type of data, accounting for different aspects of the urban environment.
\item We develop a multi-graph aggregation model to effectively integrate information from multiple graphs to capture both short-term local dependencies among geographic neighbor regions and long-term global correlations among regions with similar functions.
\item We devise an effective dual-stage fusion module that employs adaptive weights to facilitate information sharing across diverse views, incorporating attentive attention and gated fusion mechanisms. This module aims to enhance the efficiency and effectiveness of the fusion process in the urban region profiling model.
\item We extensively evaluate our approach using real-world data through a series of experiments. The results demonstrate the superiority of our method compared to state-of-the-art baselines. Additionally, our approach exhibits significantly improved computational efficiency.
\end{itemize}       

The rest of the paper is organized as follows. Section \ref{sec2} presents related work. Section \ref{sec3} gives the problem definition. Section \ref{sec4} elaborates on the proposed model. Section \ref{sec5} presents an extensive experiment evaluation. Section \ref{sec6} concludes the paper.

\section{Related Work}
\label{sec2}
This section reviews related work from three aspects: graph embedding, region representation learning, and attention mechanism.

\subsection{Graph Embedding}       
Graph embedding aims to acquire low-dimensional vectors that effectively represent vertices in a graph, preserving both the network structure and attributes. This enables the use of the embedding for diverse downstream tasks, including node classification and clustering.  Traditional approaches to graph embedding are based on adjacency matrices, Laplacian matrices, or other variants \cite{Roweis_2000_Nonlinear, Belkin_NIPS2001_Laplacian}. Motivated by word embedding models, various algorithms, including DeepWalk\cite{Perozzi_2014_DeepWalk}, LINE\cite{Tang_2015_LINE}, and Node2Vec\cite{Grover_2016_node2vec}, have been introduced. These algorithms generate random walks to capture local structures and learn vertex embeddings. However, they face challenges in capturing global topology and propagating feature information. More recently, GNNs have demonstrated promising representational capabilities, falling into two main categories: spectrum-based GNNs and space-based GNNs. Spectrum-based GNNs, exemplified by ChebNet\cite{Defferrard_2017_ChebNet} and Graph Convolution Network\cite{Kipf_2017_SemiSupervised}, execute graph convolution in the Fourier domain. Spatially based GNNs, such as Graph Isomorphism Network\cite{Xu_2019_How}, GraphSAGE\cite{Hamilton_2018_GraphSAGE}, and GAT\cite{Velickovic_2018_GAT}, perform graph convolution operations directly in the graph domain. Many graph embedding methods based on graph neural networks have thus emerged \cite{Abu-El-Haija_2018_Watch, Kipf_2016_Variational, Gilmer_2017_Neural}. Several graph embedding methods based on GNNs have emerged \cite{Abu-El-Haija_2018_Watch, Kipf_2016_Variational, Gilmer_2017_Neural}. Nevertheless, these methods face challenges in effectively capturing long-range dependencies and non-linear relationships between graph embeddings.

\subsection{Region Representation Learning}
In recent years, the surge in urban data availability has fueled research efforts focused on learning representations from urban areas. The primary objective is to learn spatial similarities, aligning regions with high similarity to geometrically proximate spaces. Numerous studies, including \cite{Jean_2018_Tile2Vec, Wang_2020_Urban2Vec, Wang_2017_HDGE, Hui_2020_Predicting, region2vec}, have delved into region embeddings for predicting region-specific features. Some approaches leverage mobility data for modeling region associations, such as those based on mutual information \cite{Yao_2018_ZeMob} or grammatical targets \cite{Zhang_2017_RPA} in human mobility. Notably, MV-PN \cite{Fu_2019_MVPN} incorporates intra-regional POI networks and spatial autocorrelation layers for learning intra and inter-regional similarities. MVGRE \cite{Zhang_2020_MVGRE} explores multi-view region similarity using taxi trip records, check-ins, and POI category distribution. While MGFN \cite{Wu_2022_MGFN} focuses on extracting traffic patterns for region representations, it overlooks crucial POI data. HREP \cite{HREP} introduces a continuous prompt method but has limitations in capturing nonlinear relationships over long distances in regions.

\subsection{Attention Mechanism}
Attention mechanisms, widely applied in diverse fields such as natural language processing, speech recognition, and image captioning \cite{Vaswani_2017_Attention, Shen_2017_DiSAN}, have more recently found applications in graphs \cite{Velickovic_2018_GAT} and urban region embeddings. For instance, MVGRE \cite{Zhang_2020_MVGRE} utilizes GAT to learn vertex representations from different graphs, employing self-attention to disseminate knowledge across representations of various views. Region2Vec \cite{region2vec} leverages GAT and Transformer to enhance the propagation and fusion of data sources from different views. MGFN \cite{Wu_2022_MGFN} introduces a multi-level cross-attention mechanism for learning integrated embeddings from multiple moving patterns based on intra and inter-pattern information. In contrast to these methods, our attention mechanism excels in efficiency and accuracy, adept at propagating knowledge of region representations over long distances and extracting profound relationships between different regions.

\section{Preliminaries}
\label{sec3}
In this section, we first give some notations and define the urban region embedding problem. Supposed that a city is partitioned into $N$ regions $V = \{v_1, v_2, \cdots , v_N\}$, where $v_i$ denotes the $i$-th region. For a \textbf{trip} $p=(v_o,v_d)$, $v_o$ denotes the origin region and $v_d$ denotes the destination region where $1\leq o,d \leq N$. Given the set of regions $V$, we further give the following definitions.
\begin{definition}[Human mobility feature]
The human mobility feature is defined as a trip sets $\mathcal{P}=\{p_i|p_i.{v_o},p_i.{v_d}\in V\}$
that occur in urban areas, $i=\{1,2,\cdots,M\}$ and $M$ is the number of trips.
\end{definition}

\begin{definition}[Function feature]
The function feature characterizes the functions of regions, such as entertainment, medical, and education. In this paper, the function feature of the region is characterized by POIs in the located region. Given a region $v_i$ that contains multiple categories of POIs, its function feature is denoted as: 

\begin{equation}
\mathcal F=\left\{f_i^k \mid f_i^k \in \mathbb{R}^C\right\}, i=\left\{1,2, \cdots, N\right\}.
\end{equation}
where $C$ is the number of POIs types in the city, $f_i^k$ denotes the number of places located in $v_i$ with the $k-$th category of POIs, $k=\{1,2,\cdots, C\}$.
\end{definition}

\begin{definition}[Semantics feature]
The semantics feature describes the activities of POIs in regions, which integrates human activities and POIs information. In particular, we use check-in information to characterize the semantics feature. Given a region $v_i$ containing $M$ check-in points, its semantics feature is given by
\begin{equation}
\mathcal S=\left\{s_i^k \mid s_i^k \in \mathbb{R}\right\}, i=\left\{1,2, \cdots, N\right\}.
\end{equation}
where $s_i^k$ denotes the number of check-in records in $v_i$ with the $k-$th check-in point, $k=\{1,2,\cdots, \mathbf{K}\}$.
\end{definition}

\begin{definition}[Urban region embedding problem]
Using the input of human mobility feature $\mathcal P$, function feature $\mathcal F$ and semantics feature $\mathcal S$, the urban region embedding problem aims to learn a low dimensional embedding of the regions in a city. This embedding can be employed for downstream tasks, such as check-in prediction, land usage classification, and crime prediction. 
\end{definition}

\section{METHODOLOGY}
\label{sec4}
The overall architecture of ATGRL proposed in this paper is shown in Figure.~\ref{architecture}, which consists of three parts: the graph-enhanced learning module, the multi-graph aggregation module, and the dual-stage fusion module.
\subsection{Overview}
Inspired by recent advances in multi-view graph neural networks for urban region representation learning, our approach produces more robust embeddings by mining the nonlinear relationships between long-range regions and adequately fusing multi-view region representations in dual stages. 

\begin{figure}[!t]
\centering\includegraphics[width=1.0\textwidth]{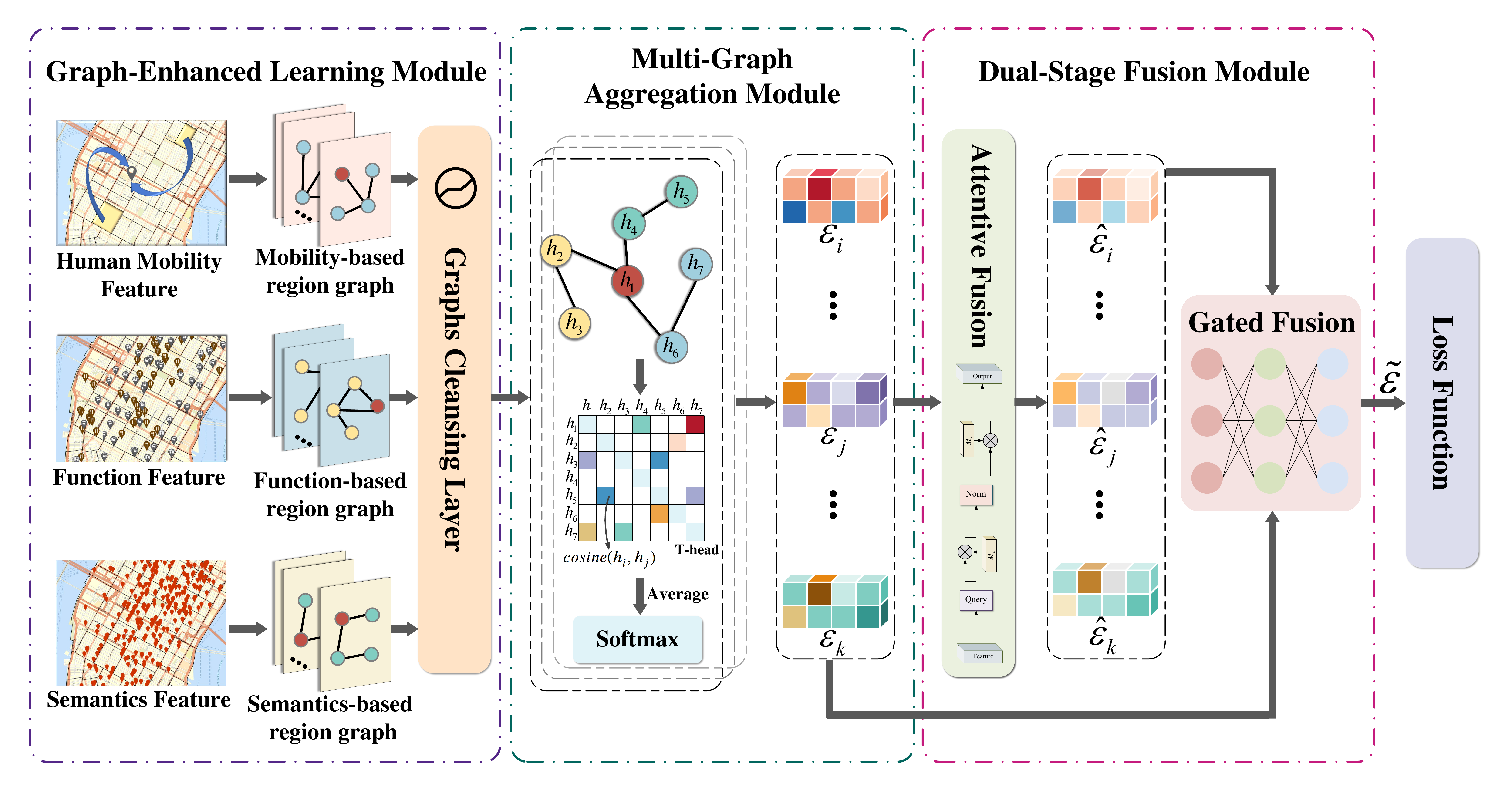}
\caption{The architecture of ATGRL consists of three components: (A) The graph-enhanced learning module for constructing multi-graph from features extracted from human mobility data, functional features, and semantic features and filtering the graphs for noise using soft thresholding. (B) The multi-graph Aggregation Module for capturing non-linear and long-range dependencies between regions using improved cosine similarity graph attention mechanism to generate corresponding region representations. (C) The dual-stage fusion module utilizes an improved linear attention mechanism to achieve information sharing between different views and efficiently fuses multi-view representations for urban region embedding.}
\label{architecture}
\end{figure}

The overall architecture of ATGRL proposed in this paper is shown in Figure.~\ref{architecture}, which consists of the following components:
\begin{itemize}
\item A region graph learning method is proposed to learn region-wise dependencies from three different types of features and represent them as multiple graphs. The graph's vertices represent regions and their edges encode pairwise relationships between regions.
\item A multi-graph aggregation method, i.e., a new graph attention, aims to extract the similar region conditions of urban networks, not just neighborhood regions.
\item A dual-stage graph fusion method, incorporating an enhanced self-attention mechanism and MLP network enables the fusion of global and local information for multiple views.
\end{itemize}
\subsection{Graph-Enhanced Learning Module}
In this section, we propose a graph-enhanced learning module to model different types of dependencies between regions. In particular, it encodes multiple graphs, including mobility-based region graphs, function-based region graphs, and semantics-based region graphs.

\subsubsection{Mobility-based region graph}
The analysis of human mobility within urban spaces, particularly across regions, is facilitated by their trajectories. When individuals travel between different origins ($O$) and destinations ($D$), patterns exhibit similarities if they share the same O/D region. By investigating the similarity of these patterns, crucial features correlated to human mobility can be identified. For a given set of human mobility feature $\mathcal P$, the similarity value between region $v_o$ and region $v_d$ can be calculated using the expression:
\begin{equation}
s_{v_j}^{v_i}=\left|\left\{\left(v_o, v_d\right) \in \mathcal{P} \mid v_o=v_i, v_d=v_j\right\}\right|,
\end{equation}
where $|.|$ denotes the size of the trip. Then we employ distributions $p_o\left(v \mid v_i\right)$ and $p_d\left(v \mid v_i\right)$ to describe the origin and destination contexts of a region $v_i$ as follows:
\begin{equation}
p_o\left(v \mid v_i\right)=\frac{s_{v_i}^v}{\sum_v s_{v_i}^v}, \quad p_d\left(v \mid v_i\right)=\frac{s_v^{v_i}}{\sum_v s_v^{v_i}}.
\end{equation}
The two types of dependencies were defined by us based on the origin and destination context of each region, as follows,
\begin{equation}
\mathcal{D}_O^{i j}=\operatorname{sim}\left(p_o\left(v \mid v_i\right), p_o\left(v \mid v_j\right)\right),
\label{DO}
\end{equation}
\begin{equation}
\mathcal{D}_D^{i j}=\operatorname{sim}\left(p_d\left(v \mid v_i\right), p_d\left(v \mid v_j\right)\right),
\label{DD}
\end{equation}
where $\mathcal{D}_O^{i j}$ denotes the dependencies between two origins, $\mathcal{D}_D^{i j}$ represents the dependencies between two destinations, $\operatorname{sim}(\cdot)$ denotes the cosine similarity. Utilizing Eqs. \ref{DO} and \ref{DD}, we proceed to construct region-wise graphs $\mathcal{G}_O$ and $\mathcal{G}_D$.

\subsubsection{Function-based region graph}
In urban environments, the functionality of regions is encapsulated by the information about PoIs. The attributes of PoIs serve as indicators of the functions specific regions fulfill. To integrate PoIs context into region embeddings, we adeptly incorporate functional information into the representation space. For a given function feature $f_i^k$, this process is expressed as follows:
\begin{equation}
\label{GF}
\mathcal{D}_{F}^{i j}=\operatorname{sim}\left({f}_i^k, {f}_j^k\right),
\end{equation}
where $\mathcal{D}_{F}^{i j}$ denotes the functional dependency between region $v_i$ and $v_j$. Consequently, we establish the function-based region graph $\mathcal{G}_F$.

\subsubsection{Semantics-based region graph}
The check-in data provides insight into human activity, reflecting the popularity of each POIs category. We utilize the semantic features of regions explored using check-in data attributes to identify the importance of each check-in type within a given region using semantic dependencies. Given the semantics feature $s_i$, the procedure can be formulated as follows:
\begin{equation}
\label{GS}
\mathcal{D}_{S}^{i j}=\operatorname{sim}\left({s}_i^k, {s}_j^k\right)
\end{equation}

where $\mathcal{D}_{S}^{i j}$ is the semantics dependency between region $v_i$ and $v_j$. 
Consequently, we establish the semantics region graph $\mathcal{G}_{S}$.
\subsubsection{Graph Cleansing Layer}
To reduce the impact of data noise and incompleteness in complex urban environments and data acquisition processes without increasing the time-consuming burden, we employ a lightweight denoising method, the soft-thresholding method, which is commonly used for various signal denoising and feature compression\cite{Donoho_1995_Denoising, Isogawa_2018_Shrinkage, Zhao_2020_Residual, WANG2021512soft}. Deep learning empowers soft thresholding as an effective method for suppressing noise-related information and improving feature modeling performance. In urban region embedding, different urban regions have different degrees of importance due to their unique features. The application of the soft thresholding technique can zero out certain less important features in the embedding representation, thus improving the sparsity of the region embedding. Meanwhile, soft thresholding is an element-wise noise reduction method with relatively low computational complexity, which optimizes the use of storage and computational resources. Specially, the function of soft thresholding is defined as:
\begin{equation}
\label{noise}
y=\left\{\begin{array}{cl}
x-\tau & x>\tau \\
0 & -\tau \leq x \leq \tau \\
x+\tau & x<-\tau
\end{array}\right.
\end{equation}
where $x$ is the input feature, $y$ is the output feature and $\tau$ is the threshold value, i.e. the positive parameter. 

\begin{figure}[!t]
\centering
\subfigure[]{\includegraphics[scale=.3]{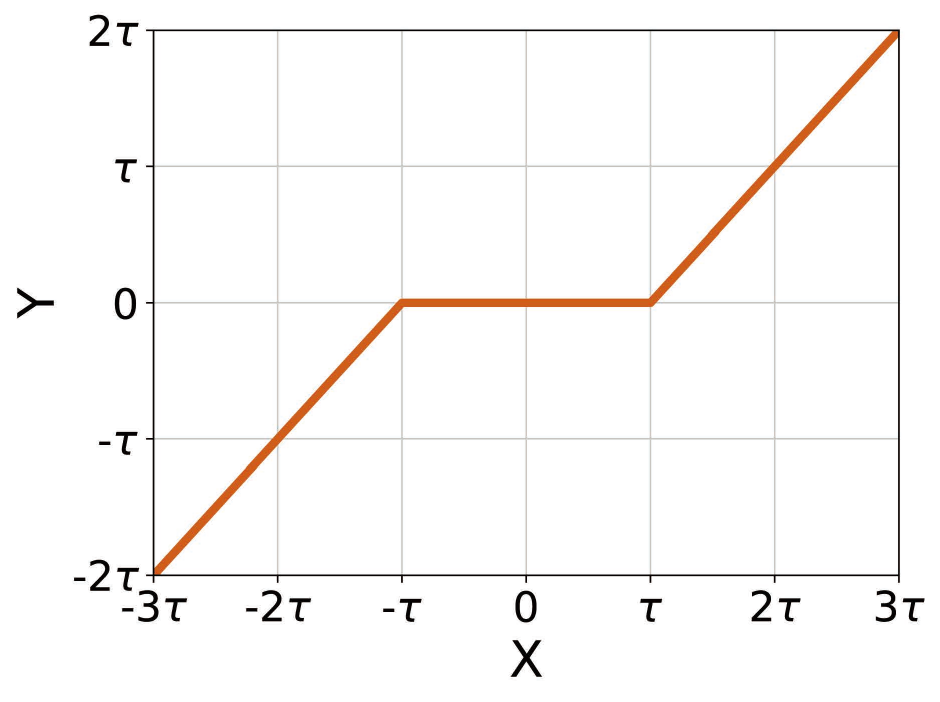}
\label{soft}}
\subfigure[]{\includegraphics[scale=.3]{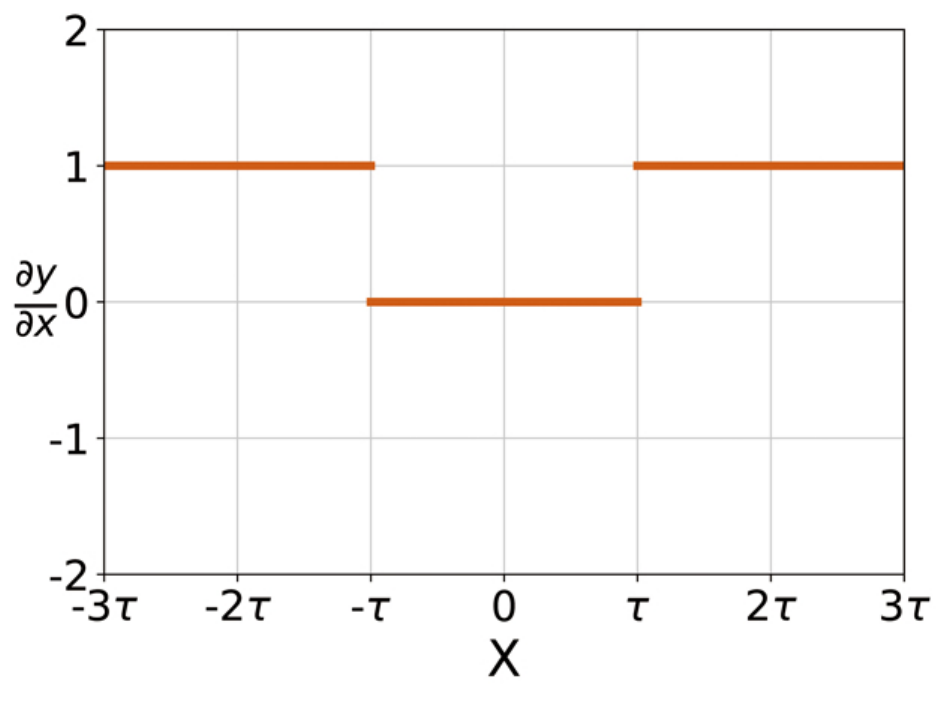}
\label{deriv}}
\caption{Illustration of (a) soft thresholding and (b) its derivative.}
\label{SL}
\end{figure}

In the proposed model, we employ soft thresholding in the regions-wise graph learning model. As shown in Figure.~\ref{soft}, instead of setting negative features to 0 in the ReLU activation function, soft thresholding sets features close to 0 as 0 and preserves the useful negative features. Moreover, as depicted in Figure.~\ref{deriv}, soft thresholding is effective in mitigating gradient vanishing and exploding issues, as the derivative of the output concerning the input is either 1 or 0. The derivative can be mathematically expressed as follows:
\begin{equation}
\frac{\partial y}{\partial x}=\left\{\begin{array}{ll}
1 & x>\tau \\
0 & -\tau \leq x \leq \tau \\
1 & x<-\tau
\end{array} .\right.
\end{equation}

In this study, we set the initial threshold to 0. To dynamically adjust the threshold for optimal model performance, we employ the gradient descent optimization algorithm. The update rule is given by:
\begin{equation}
\lambda_{\text {}}=\lambda_{\text {}}-\eta \cdot \nabla_\lambda \mathcal{L},
\end{equation}
which $\eta$ is the learning rate, $\mathcal{L}$ is the weight and $\nabla_\lambda$ is the gradient with respect to $\lambda$.

With the help of the graphs cleansing layer, the reduction of noise and irrelevant information in the feature matrix improves the quality of the urban region embedding and also increases the generality of the model. As a result, we can get denoised region graphs $\mathcal{G}_O^{\prime}$, $\mathcal{G}_D^{\prime}$, $\mathcal{G}_F^{\prime}$, and $\mathcal{G}_S^{\prime}$.

\subsection{Multi-Graph Aggregation Module}
It is observed that not only adjacent regions are relevant, but many long-range PoIs also have regional correlations. 
Nonetheless, existing works such as \cite{Zhang_2020_MVGRE, Luo_2022_ProfilingFramework} use the GAT mechanism to assign weights based on the similarity of different neighboring nodes, which can lead to an inability to capture the long-distance dependencies and non-linear relationships between regions in urban graph networks. 

To model the correlations of any two regions in the whole urban network, we first need to get the similarity measurement between each region pair. Dot product and cosine similarity are usually used as the similarity functions. 
In this article, we take the cosine similarity as an example. Given the vertex feature $\mathbf{h}=\left\{{h}_1, {h}_2, \ldots, {h}_N\right\}, {h}_i \in \mathbb{R}^C$, where $N$ is the number of nodes, and $C$ is the input dimension, the multi-graph aggregation module works as follows:
    \begin{equation}\label{cosAtt}
     \begin{split}
        &A_{ij}=cosine(h_i,h_j)\cdot w_{ij}=\frac{h_i(h_j)^T\cdot w_{ij}}{\lVert h_i\rVert \lVert h_j\rVert},
     \end{split}
    \end{equation}
where $\lVert h_i\rVert$ denotes the norm of vector $h_i$ and $\cdot$ is the dot product of vectors. $w_{ij}$ is the weight matrix and $A_{ij}$ is the similarity between $h_i$ and $h_j$. 
    
Next, $A_{ij}$ is utilized to aggregate information from all other features in the network to each feature, 
    
\begin{equation}\label{cosAtt1}
        \hat{h_i}=\sigma\left(\sum_{j\in \widetilde{N}_i}A_{ij}h_jw_{ij}\right),
    \end{equation}
where $\widetilde{N}_i$ is nodes set in the graph except node $h_i$. $\sigma$ denotes the softmax function. $\hat{h_i}$ is the aggregation information from the global features to the feature $h_i$. 
    
To further improve performance, we incorporate the multi-head attention mechanism into the multi-graph aggregation procedure as\cite{Velickovic_2018_GAT}. Specifically, we adopt $T$ independent attention mechanisms to perform the transformation of Eq.\ref{cosAtt1} and then we average their features to obtain the following output feature representation:
    \begin{equation}
    \label{THead}
    \hat{h_i}=\sigma\left(\frac{1}{T} \sum_{t=1}^T \sum_{j \in \mathcal{N}_i} A_{ij}^t {h}_j w_{ij}^t\right)
    \end{equation}
where $A_{ij}^t$ are normalized attention coefficients computed by the $t$-th attention mechanism, and $w_{ij}^t$ is the corresponding weight matrix.

Consequently, the output of the multi-graph aggregation module is denoted as $\mathcal{E}_O$, $\mathcal{E}_D$, $\mathcal{E}_F$, and $\mathcal{E}_S$, which are corresponding to the representation results for three types graph.

\subsection{Dual-Stage Fusion Module}
While self-attention has proven effective in graph embedding learning, its computational costs are noteworthy. To address this issue, we propose a linear attention mechanism to alleviate the computational burden in urban region embedding. This mechanism is integrated into a gated fusion module, forming a dual-stage fusion module for efficient global view fusion.

\subsubsection{Attentive Fusion}
In various fields such as estate price valuation, crime prediction, and check-in prediction, strong correlations have been observed between different regions. For example, studies have found that housing prices in different areas are closely related to factors such as transportation, education, commercial and natural resources. Similarly, in crime prediction, certain areas may have a higher crime rate due to factors such as unemployment, poverty, and education levels in the surrounding area. In terms of check-in point prediction, there is also a strong correlation between different regions. For instance, commercial check-in points in certain areas may be related to factors such as traffic, population density, and consumption levels in the surrounding area. These observations suggest that combining information from multiple perspectives can enhance the learning process of individual views and improve the representation performance.

Recently, self-attention has become prevalent in various fusion methods\cite{Zhang_2020_MVGRE, Wu_2022_MGFN}. However, it has the disadvantage of high computational cost. For instance, given a region embedding $\mathcal{E} \in \mathbb{R}^{N \times d}$ where $N$ is the number of regions and $d$ is the number of feature dimensions, self-attention linearly projects the input to a query matrix $Q \in \mathbb{R}^{N \times d^{\prime}}$, a key matrix $K \in \mathbb{R}^{N \times d^{\prime}}$, and a value matrix $V \in \mathbb{R}^{N \times d}$. Then self-attention can be formulated as: 
\begin{equation}
A=(\alpha)_{i, j}=\operatorname{softmax}\left(Q K^T\right)
\end{equation}

\begin{equation}
F_{\text {out }}=A V,
\end{equation}
where $A \in \mathbb{R}^{N \times N}$ is the attention matrix and $\alpha_{i, j}$ is the pair-wise affinity between the $i$-th and $j$-th elements. Thus, the high computational complexity of $O\left(d N^2\right)$ brings significant drawbacks to the use of self-attention. 
To reduce the computational complexity, we design a linear attention to support the propagation of information across multiple views.
Given the representations of $M$ views $\{\mathcal{E}_1,\mathcal{E}_2,\cdots,\mathcal{E}_M\}$, for each $\mathcal{E}_i$, we then propagate information among all views as follows:
\begin{equation}
\left[A_i\right]_{i=1}^M=\operatorname{Norm}\left(\left[{\mathcal{E}_i M_k^T}\right]_{i=1}^M\right),  \hat{\mathcal{E}}_i=\sum_{i=1}^M A_i M_v\label{linearAtt}
\end{equation}
where \(M_k \in \mathbb{R}^{H \times d}\) and \(M_v \in \mathbb{R}^{H \times d}\) are trainable parameters independent of the representations. They serve as the key and value memory for the entire training dataset. Note that the value of \(d\) is equivalent to the dimension of the model. Since $d$ and $H$ are hyper-parameters and the computational complexity of linear attention is $O\left(dHN\right)$, the proposed algorithm is linear in the number of regions. $\hat{\mathcal{E}}_i$ is considered as the relevant global information for $i$-th view. When multiple sources of data are fused, information bottlenecks are inevitable due to the information compression. To reduce the influence of noise from the global fusion process, we employ the softmax operator and an $l_1$-norm to normalize\cite{Guo_2021_PCT}. 
The embedding results, denoted as $\hat{\mathcal{E}}_O$, $\hat{\mathcal{E}}_D$, $\hat{\mathcal{E}}_F$, and $\hat{\mathcal{E}}_S$, are obtained from the previously described modules. Additionally, the global information extracted is further fed into subsequent gated fusions within the model. This iterative process allows the model to refine its understanding by incorporating both local and global perspectives, contributing to a more comprehensive and refined representation of urban regions.

\subsubsection{Gated Fusion}
This module is designed to effectively integrate global and local region representation. The fusion layer works in the following way:
\begin{equation}
\mathcal{E}_i^{\prime}=\mathbf{a}_i \hat{\mathcal{E}}_i+(1-\mathbf{a})_i \mathcal{E}_i
\end{equation}

\begin{equation}
\mathcal{E}_{\mathcal{F}}=\sum_i^M w_i \mathcal{E}_i, \quad w_i=\sigma\left(\mathcal{E}_i W_f+b_f\right)
\end{equation}
where $w_i$ represents the weight of the $i$-th view, learned through a single-layer MLP network with the $i$-th embeddings as input. $\mathcal{E}_i^{\prime}$ is the representation for the $i$-th view with global information, and $\mathbf{a}_i$ denotes the learning parameters.

To facilitate the learning process of the multi-view fusion layer, we involve $\mathcal{E}$ in the learning objective of each view. Specifically, we update the representation of each view as:
\begin{equation}
\label{Beta}
\tilde{\mathcal{E}}_i=\beta\mathcal{E}_i^{\prime}+(1-\beta) \mathcal{E_F},
\end{equation}
where $\beta$ is a hyper-parameter. By integrating the outputs of the multi-graph aggregation model into the dual-stage fusion module, we obtain region embeddings $\tilde{\mathcal{E}}_O, \tilde{\mathcal{E}}_D, \tilde{\mathcal{E}}_F$, 
and $\tilde{\mathcal{E}}_S$, which are utilized in various learning objectives.

\begin{algorithm}
\caption{ATGRL Learning Algorithm}\label{alg:model}
\begin{algorithmic}[1]
\Require Human mobility feature $\mathcal{P}$, function feature $\mathcal{F}$, semantics feature $\mathcal{S}$, maximum epoch number $E$, learning rate $\eta$
\Ensure Task region embeddings $\tilde{\mathcal{E}}$
\State Initialize all parameters
\For{$e = 1$ to $E$}
    \State Construct region graphs $\mathcal{G}_O$, $\mathcal{G}_D$, $\mathcal{G}_F$, and $\mathcal{G}_S$ using Eqs. \ref{DO}, \ref{DD}, \ref{GF}, and \ref{GS}
    \State Cleanse the region graphs using Eq. \ref{noise}
    \State Obtain region embeddings $\mathcal{E}_O$, $\mathcal{E}_D$, $\mathcal{E}_F$, and $\mathcal{E}_S$ via multi-graph aggregation (Eq. \ref{THead})
    \State Perform dual-stage fusion to get task region embeddings $\tilde{\mathcal{E}}_O, \tilde{\mathcal{E}}_D, \tilde{\mathcal{E}}_F$, and $\tilde{\mathcal{E}}_S$ (Eqs. \ref{linearAtt}, \ref{Beta})
    \State Optimize the model using loss $\mathcal{L}$ calculated by Eq. \ref{Ltotal}
\EndFor
\State \Return Task region embeddings $\tilde{\mathcal{E}}_O, \tilde{\mathcal{E}}_D, \tilde{\mathcal{E}}_F$, and $\tilde{\mathcal{E}}_S$
\end{algorithmic}
\end{algorithm}

\subsection{Loss Function}
The output of the dual-stage fusion module is the low-dimensional embeddings of all regions in the city, which are the encoded representations of each region and can be employed for the subsequent classification or prediction tasks, such as check-in prediction, land usage classification, and crime prediction. The overall objectives of the three tasks can be expressed as:
\begin{equation}
\label{Ltotal}
\mathcal{L}=\mathcal{L}_{ODP}+\mathcal{L}_{FP}+\mathcal{L}_{SP}.
\end{equation}
where $\mathcal{L}_{ODP}$, $\mathcal{L}_{FP}$, and $\mathcal{L}_{SP}$ are losses for origin and destination predictor, functional predictor, and semantics predictor respectively. To facilitate a clear presentation of our proposed algorithm, we employ pseudo-code to delineate its principal steps, as shown in Algorithm \ref{alg:model}.
\subsubsection{Origin and destination predictor}
Through the origin and destination predictor, we can predict the destination region when the origin region is provided, or conversely predict the origin region given the destination region. In this context, the regions are represented as $\tilde{\mathcal{E}}_O=\left\{e_o^i\right\}_{i=1}^n$ and $\tilde{\mathcal{E}}_D=\left\{e_d^i\right\}_{i=1}^n$. Given an origin region $v_i$, we formulate the distribution of the target region $v_j$ as follows:
\begin{equation}
\hat{p}_O\left(v_j \mid v_i\right)=\frac{\exp \left(e_O^{i T} e_D^j\right)}{\sum_j \exp \left(e_O^i e_D^j\right)}.
\end{equation}
Consequently, for given a destination region $v_i$, we formulate the distribution of the origin region $v_j$ as follows:
\begin{equation}
\hat{p}_D\left(v_j \mid v_i\right)=\frac{\exp \left(e_D^{i^T} e_O^j\right)}{\sum_j \exp \left(e_D^{i^T} e_O^j\right)}.
\end{equation}
Next, we consider a human mobility dataset $\mathcal{P}$ comprising actual pairs of origin and destination regions. To train our predictor, we define the loss function as the negative log-likelihood of the predicted distribution. This can be expressed as:
\begin{equation}
\mathcal{L}_{ODP}=\sum_{\left(v_i, v_j\right) \in \mathcal{P}}-\log \hat{p}_O\left(v_j \mid v_i\right)-\log \hat{p}_D\left(v_i \mid v_j\right).
\end{equation}

\subsubsection{Function predictor}
The purpose of the function predictor is to ensure that the final region representation retains the information from the function features. To achieve this, we define the prediction loss function based on $\mathcal{C}_{G}$ and $\tilde{\mathcal{E}}_{G}=\left\{e_{G}^i\right\}_{i=1}^n$ as follows:
\begin{equation}
\mathcal{L}_{FP}=\sum_{i, j}\left(\mathcal{C}_{G}^{i j}-e_{G}^{i{ }^T} e_{G}^j\right)^2.
\end{equation}

\subsubsection{Semantics predictor}
Likewise, the loss function for the semantics predictor is based on $\mathcal{C}_{S}$ and $\tilde{\mathcal{E}}_{S}=\left\{e_{S}^i\right\}_{i=1}^n$, as illustrated below:
\begin{equation}
\mathcal{L}_{SP}=\sum_{i, j}\left(\mathcal{C}_{S}^{i j}-e_{S}^{i{ }^T} e_{S}^j\right)^2.
\end{equation}

\section{EXPERIMENTS}
\label{sec5}
In this section, extensive experiments are conducted to verify the superiority of the proposed model.
\begin{table}[!t]
\caption{The details for the dataset.}
\centering
\begin{tabular}{@{}cc@{}}
\toprule
Dataset       & Details                                              \\ \midrule
Regions       & 180 regions based on the Manhattan community boards. \\
Taxi trips    & 10 million taxicab trips during one month.           \\
Check-in data & 100,000 check-in points with 200 categories.         \\
POI data      & 20,000 PoI in 13 categories in the studied areas.    \\
Crime data    & 40,000 criminal records in a year                    \\ \bottomrule
\end{tabular}
\label{table1}
\end{table}

\begin{table}[!t]
\caption{Performance comparison of different approaches for check-in prediction, land usage classification, and crime prediction tasks.}
\centering
\begin{tabular}{@{}ccccccccc@{}}
\toprule
\multirow{2}{*}{Models} & \multicolumn{3}{c}{Check-in Prediction} & \multicolumn{2}{c}{Land Usage Classification} & \multicolumn{3}{c}{Crime Prediction} \\ \cmidrule(l){2-9} 
            & MAE    & RMSE   & $R^2$ & NMI  & ARI  & MAE    & RMSE   & $R^2$ \\ \midrule
LINE        & 564.59 & 853.82 & 0.08  & 0.17 & 0.01 & 117.53 & 152.43 & 0.06  \\
node2vec    & 372.83 & 609.47 & 0.44  & 0.58 & 0.35 & 75.09  & 104.97 & 0.49  \\
HDGE        & 399.28 & 536.27 & 0.57  & 0.59 & 0.29 & 72.65  & 96.36  & 0.58  \\
ZE-Mob      & 360.71 & 592.92 & 0.47  & 0.61 & 0.39 & 101.98 & 132.16 & 0.20  \\
MV-PN       & 476.14 & 784.25 & 0.08  & 0.38 & 0.16 & 92.30  & 123.96 & 0.30  \\
CGAL        & 315.58 & 524.98 & 0.59  & 0.69 & 0.45 & 69.59  & 93.49  & 0.60  \\
MVGRE       & 297.72 & 495.27 & 0.63  & 0.78 & 0.59 & 65.16  & 88.19  & 0.64  \\
MGFN        & 280.91 & 436.58 & 0.72  & 0.76 & 0.58 & 59.45  & 77.60  & 0.72  \\
ROMER       & 252.14 & 413.96 & 0.74  & 0.81 & 0.68 & 64.47  & 85.46  & 0.72  \\
HREP        & 270.28 & 406.53 & 0.75  & 0.80 & 0.65 & 65.66  & 84.59  & 0.68  \\
ATGRL(Ours) & \textbf{251.70} &\textbf{405.28} & \textbf{0.76} &\textbf{0.82}& \textbf{0.69}& 62.43       & 83.19      & 0.73     \\  \bottomrule
\end{tabular}

\label{metr}
\end{table}

\subsection{Datasets}
Experiments are conducted on several real-world datasets of New York City from the NYC open data website\footnote{https://opendata.cityofnewyork.us/}. We have chosen the borough of Manhattan in New York City as our study area, which serves as a benchmark dataset in the field of research. The study area is divided into 180 regions for comprehensive analysis. We have obtained authentic and diverse datasets, such as census block shapefiles, taxi trip data, POI data, and check-in data, from the renowned NYC Open Data platform for our research efforts. The description of each dataset can be found in Table.~\ref{table1}.

\subsection{Experimental Settings}
ATGRL has been implemented based on the TensorFlow framework. All the experiments have been carried out on an 
Intel(R) Xeon(R) CPU E5-2683 v4 hardware platform equipped with NVIDIA GeForce RTX 
3090-24G. Specifically, the model is optimized using Adam with a learning rate $\eta$ of 0.005.
For this implementation, the dimension of our model is 144, denoted as \(K=144\). The weight decay of the \(L_2\) loss on the embedding matrix is set to 0.001. The number of heads \(T\) in the linear attention mechanism is configured as 12. Concerning the hyperparameter settings, the memory unit's hyperparameters in the attentive fusion mechanism are defined as \(d=144\) and \(H=32\), and \(\beta\) is introduced with a value of 0.5 specifically for the gated fusion mechanism.

A common classification can apply to regions with congruent types of land usage classification. We employ K-means to cluster the regional embeddings that incorporate land use information from multiple graphs and present the outcomes visually for intuitive interpretation. Figure.~\ref{landuse:1} illustrates that the community boards \cite{Berg__New} partition the Manhattan administrative district into 12 districts based on land use. Accordingly, we split the study area into 12 clusters as well. The clustering outcomes should group regions with identical land use types. To further measure the regional clustering results of the embedding methods and baselines quantitatively, inspired by \cite{Yao_2018_ZeMob}, we use normalized mutual information (NMI) and adjusted Rand index (ARI).

For regression tasks (i.e., crime, check-in), we apply the Lasso regression \cite{Tibshirani_1996_Regression} with metrics of Mean Absolute Error (MAE), Root Mean Square Error (RMSE) and coefficient of determination ($R^2$) to measure the performance the models. We select $l$1 normalized weight values in the Lasso model for each method by grid searching and calculate all the metrics by K-fold cross-validation, where $K$=5.

\subsection{Baselines}
This paper compares the ATGRL model with the following baselines.

\begin{itemize}
\item {\bfseries LINE}\cite{Tang_2015_LINE}:
It is applied to preserve both the local and global network structures while optimizing the objective function.
\item {\bfseries node2vec}\cite{Grover_2016_node2vec}:
It is applied to concatenate the embedding of each graph to obtain the region embedding.
\item {\bfseries HDGE} \cite{Wang_2017_HDGE}:
It is applied to traffic flow graphs and spatial graphs for path sampling to jointly learn region representations.
\item {\bfseries ZE-Mob}\cite{Yao_2018_ZeMob} :
It is applied to the learning of region embeddings by taking into account the co-currency relation of the regions in human mobility trips. 
\item {\bfseries MV-PN}\cite{Fu_2019_MVPN} :
It is used to learn region embeddings with a region-wise multi-view POI network. We denote it as MV-PN.
\item {\bfseries CGAL}\cite{CGAL} :
It is used to encode region embeddings utilizing unsupervised methods on both the constructed POI and mobility graphs. The adversarial learning is adopted to integrate the intra-region structures and inter-region dependencies.
\item {\bfseries MVGRE}\cite{Zhang_2020_MVGRE} :
It is used to implement cross-view information sharing, and weighted multi-view fusion to extract region embeddings based on both the mobility of people and the inherent properties of regions (e.g. POI, check-in).
\item {\bfseries MGFN} \cite{Wu_2022_MGFN}:
It utilizes multi-graph fusion networks (MGFN) with a multi-level cross-attention mechanism to learn comprehensive region embeddings from multiple mobility patterns, integrated via a mobility graph fusion module.
\item {\bfseries ROMER} \cite{ROMER}:
It excels in urban region embedding by capturing multi-view dependencies from diverse data sources, employing global graph attention networks, and incorporating a dual-stage fusion module. 
\item {\bfseries HREP} \cite{HREP}:
It applies the continuous prompt method prefix-tuning to replace the direct use of region embedding in the downstream tasks, which can have different guiding effects in different downstream tasks and therefore can achieve better performance.
\end{itemize}

\subsection{Experimental Results}
\subsubsection{Main Results}
Table~\ref{metr} reports the comparison results of the crime prediction task, the land use classification task, and the check-in statistics prediction task. We observe that: 
\begin{itemize}
    \item Our method (ATGRL) outperforms all baseline tasks in the check-in prediction and land usage classification tasks. In particular, ATGRL achieves over 6.8\% improvement in MAE in the check-in prediction task. In the land classification task, more than 2.5\% improvement in ARI and more than 6\% improvement in NMI are achieved. 
   \item Traditional graph embedding methods (i.e. LINE, node2vec) perform poorly because of the local sampling approach, which can not fully express the relationships between nodes.
   \item The deep learning methods, including HDGE, ZE-Mob, MV-PN, and CGAL outperform traditional graph embedding methods. These models leverage multi-scale graph structures and embedding techniques to capture multi-level features and intricate relationships within urban areas. However, a limitation of these models lies in their neglect of the varying importance of nodes when modeling dependencies, which can impact their overall performance.
    \item Graph representation methods like MVGRE, MGFN, and HREP exhibit superior performance compared to alternative approaches. These models leverage multi-view fusion and attention mechanisms, contributing to enhanced performance. However, they all neglect long-range dependencies among different regions. Additionally, they do not account for the impact of data noise on modeling dependencies between regions, resulting in suboptimal performance.
\end{itemize}
\begin{figure}[!t]
\centering
\subfigure[MAE and RMSE in check-in prediction.]{
\includegraphics[scale=.35]{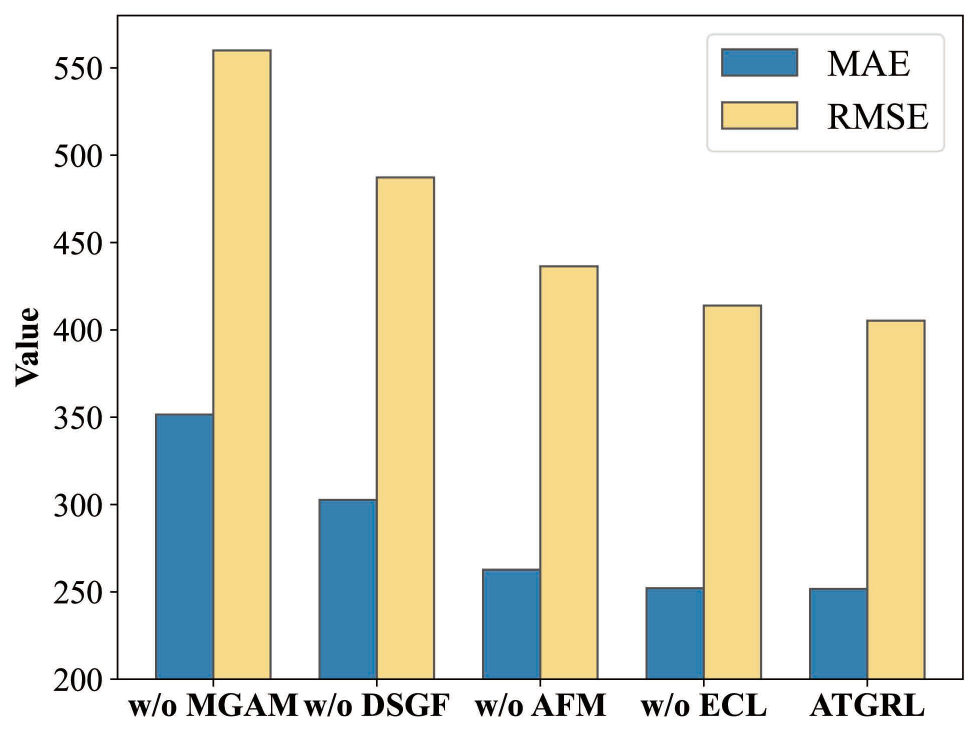}
\label{abl_chk-in}
}
\subfigure[ARI and NMI in land usage classification.]{
\includegraphics[scale=.35]{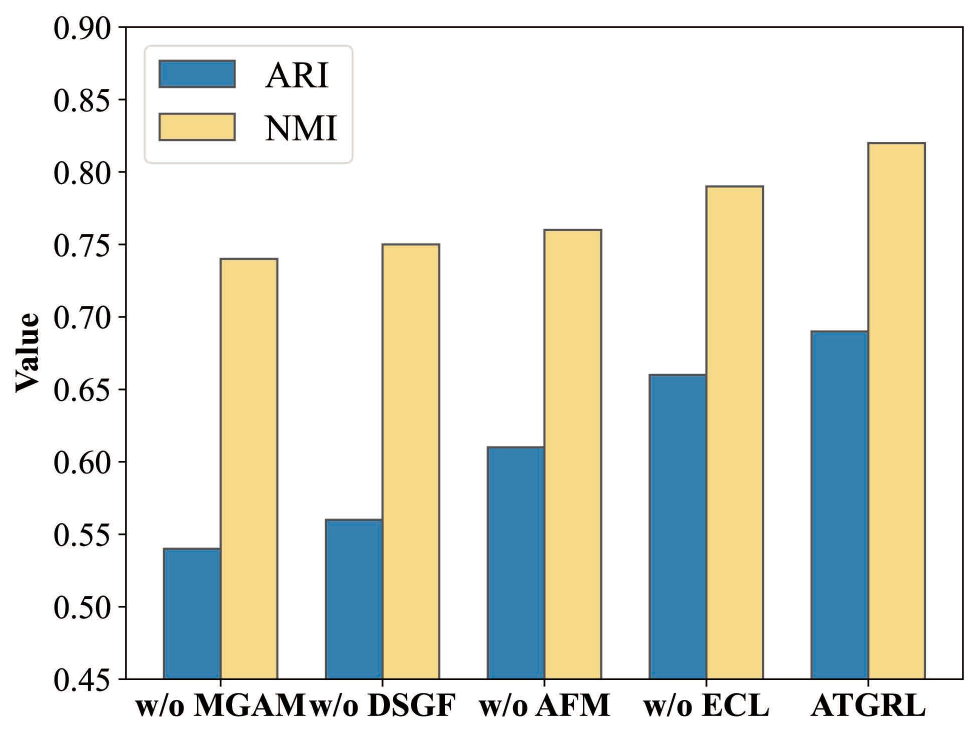}
\label{abl_landuse}
}
\caption{Ablation studies for two tasks on NYC dataset.}
\label{abla}
\end{figure}
\subsubsection{Ablation Study}
To assess the impact of key components on the proposed model, this study conducts an ablation study in land usage classification and check-in prediction tasks, respectively. The variants of ATGRL are labeled as follows:

\begin{itemize}
\item w/o GCL: It is ATGRL without graphs cleansing layer.
\item w/o MGAM: It is ATGRL without the multi-graph aggregation module, replaced with the \textbf{GAT}\cite{Velickovic_2018_GAT}.
\item w/o AFM: It is ATGRL without attentive fusion module, replaced with the \textbf{self-attention}\cite{Vaswani_2017_Attention}.
\item w/o DSGF: It is ATGRL without the dual-stage graph fusion module. The extracted spatial features are concatenated directly.
\end{itemize}

The experimental results in check-in prediction and land use classification tasks are shown in Figure.~\ref{abla}, which highlight the effectiveness of key components in ATGRL. These components include the graph-enhanced learning module, constructing graphs from various features with noise filtering, the multi-graph aggregation module, capturing nonlinear dependencies between regions using an improved cosine similarity graph attention mechanism, and the dual-stage fusion module, efficiently combining multi-view representations for urban region embedding with an improved linear attention mechanism.

\begin{figure}[!t]
\centering
\subfigure[Districts]{
\includegraphics[scale=.11]{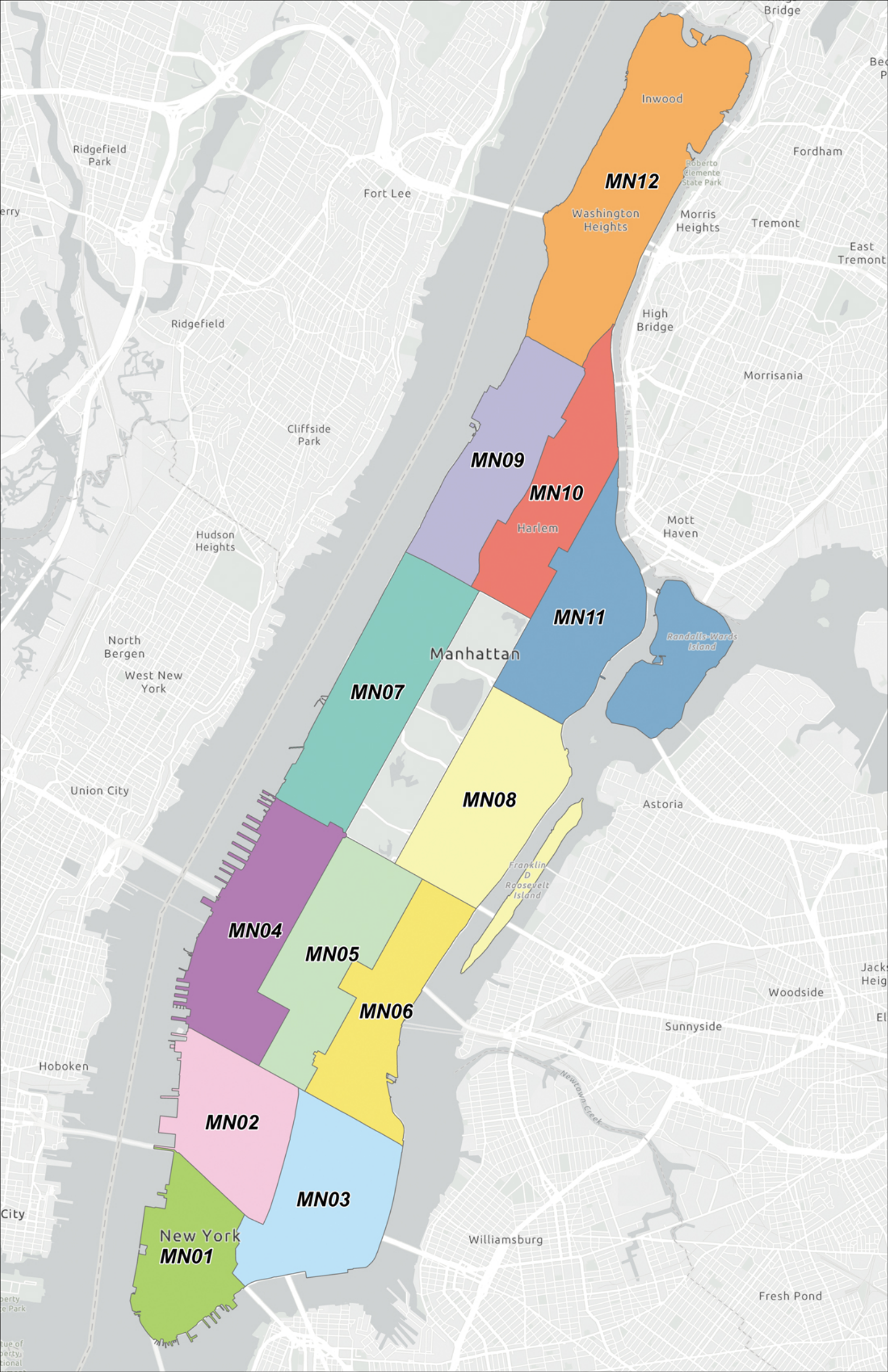} \label{landuse:1}
}
\subfigure[HDGE]{
\includegraphics[scale=.11]{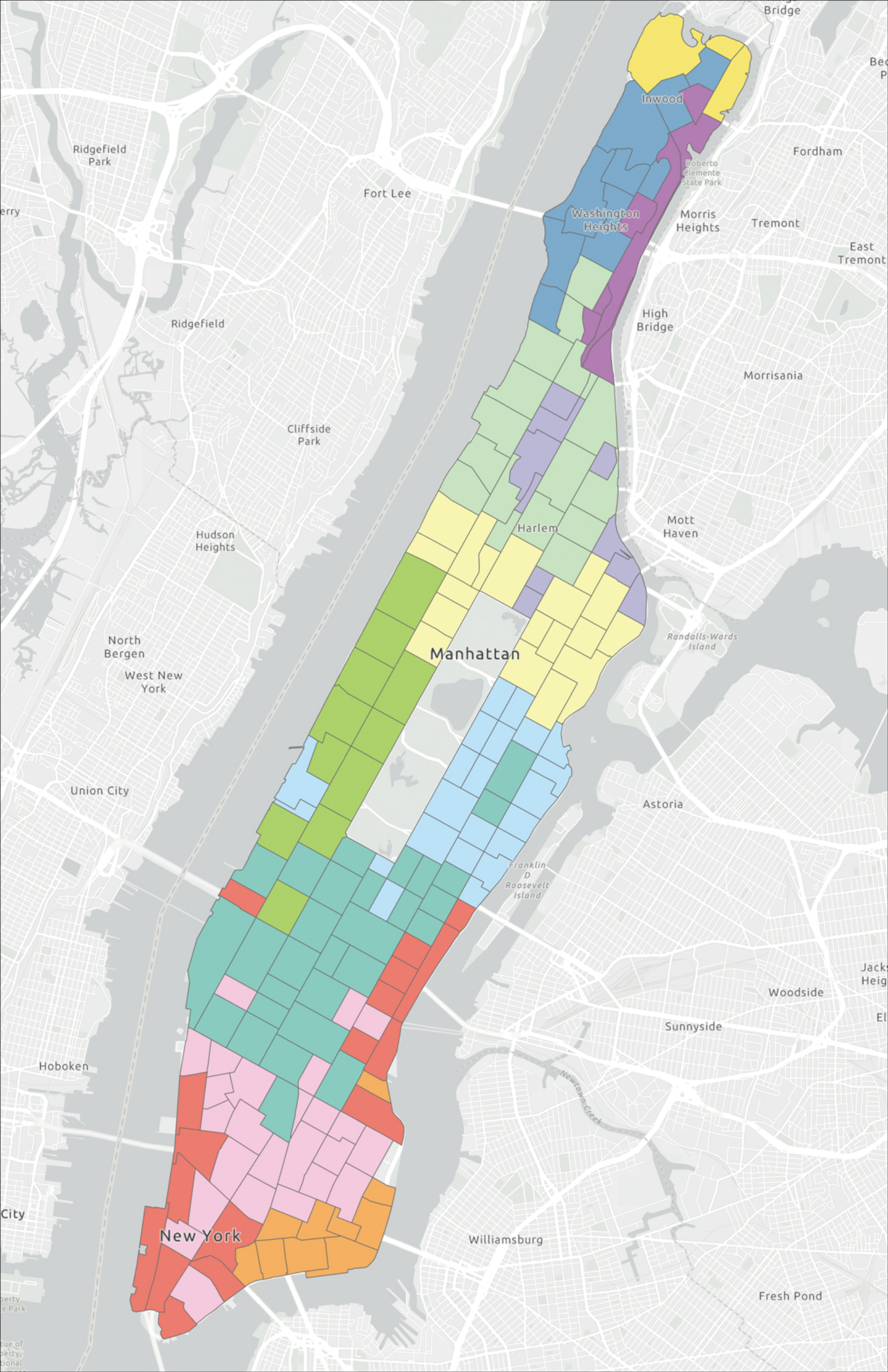}
}
\subfigure[ZE-Mob]{
\includegraphics[scale=.11]{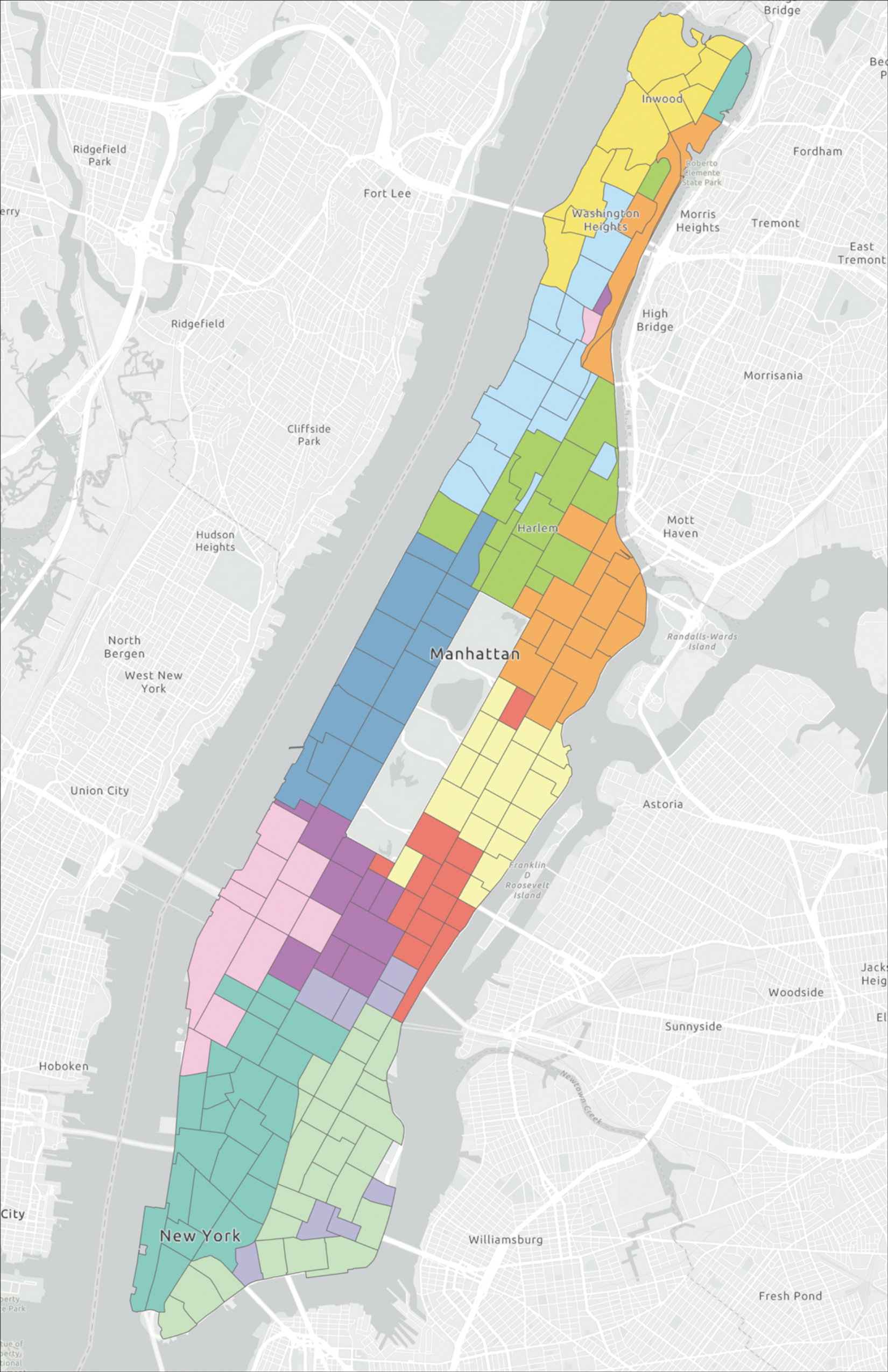}
}
\\
\subfigure[MVGRE]{
\includegraphics[scale=.11]{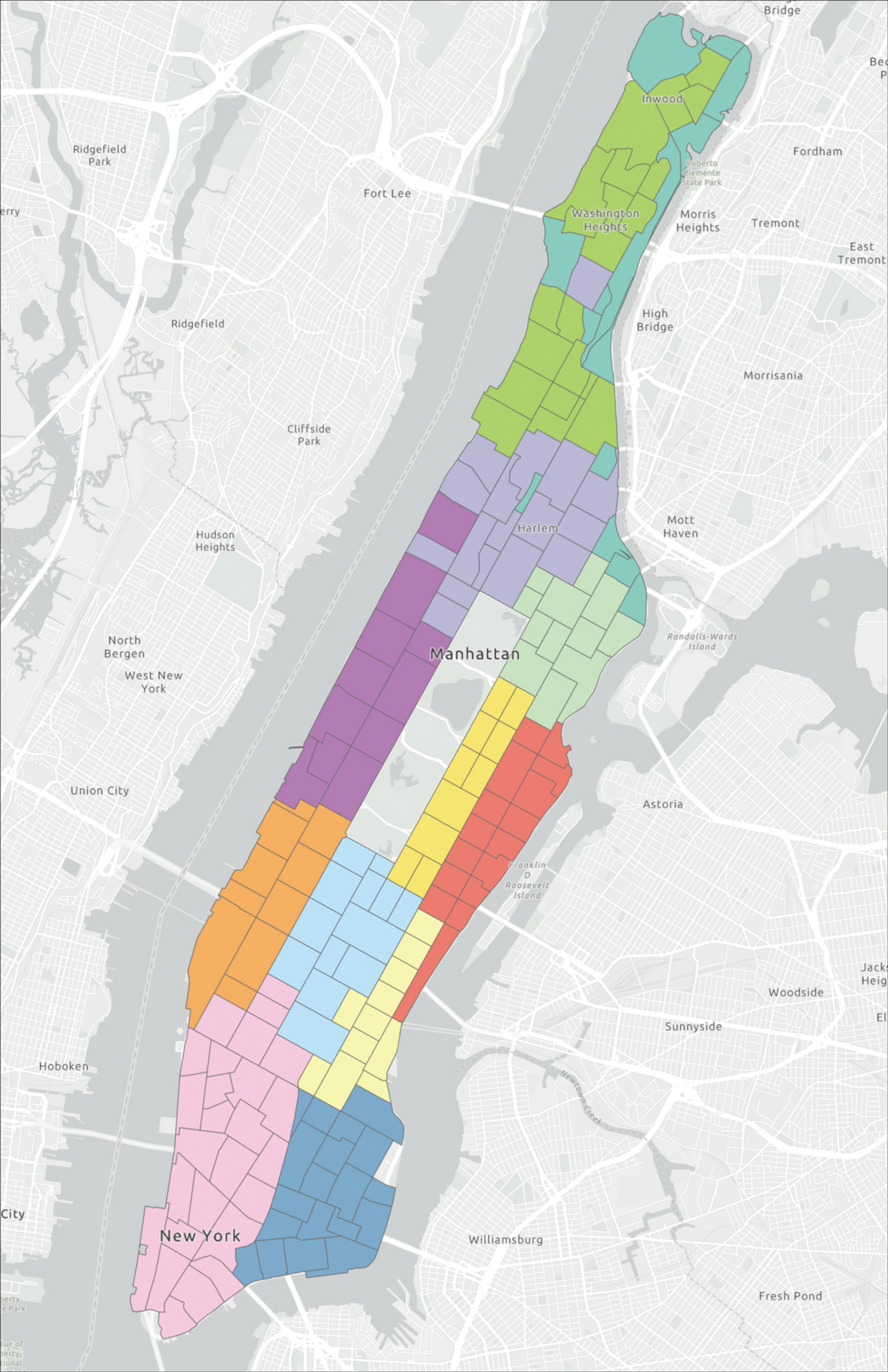}
}
\subfigure[ROMER]{
\includegraphics[scale=.11]{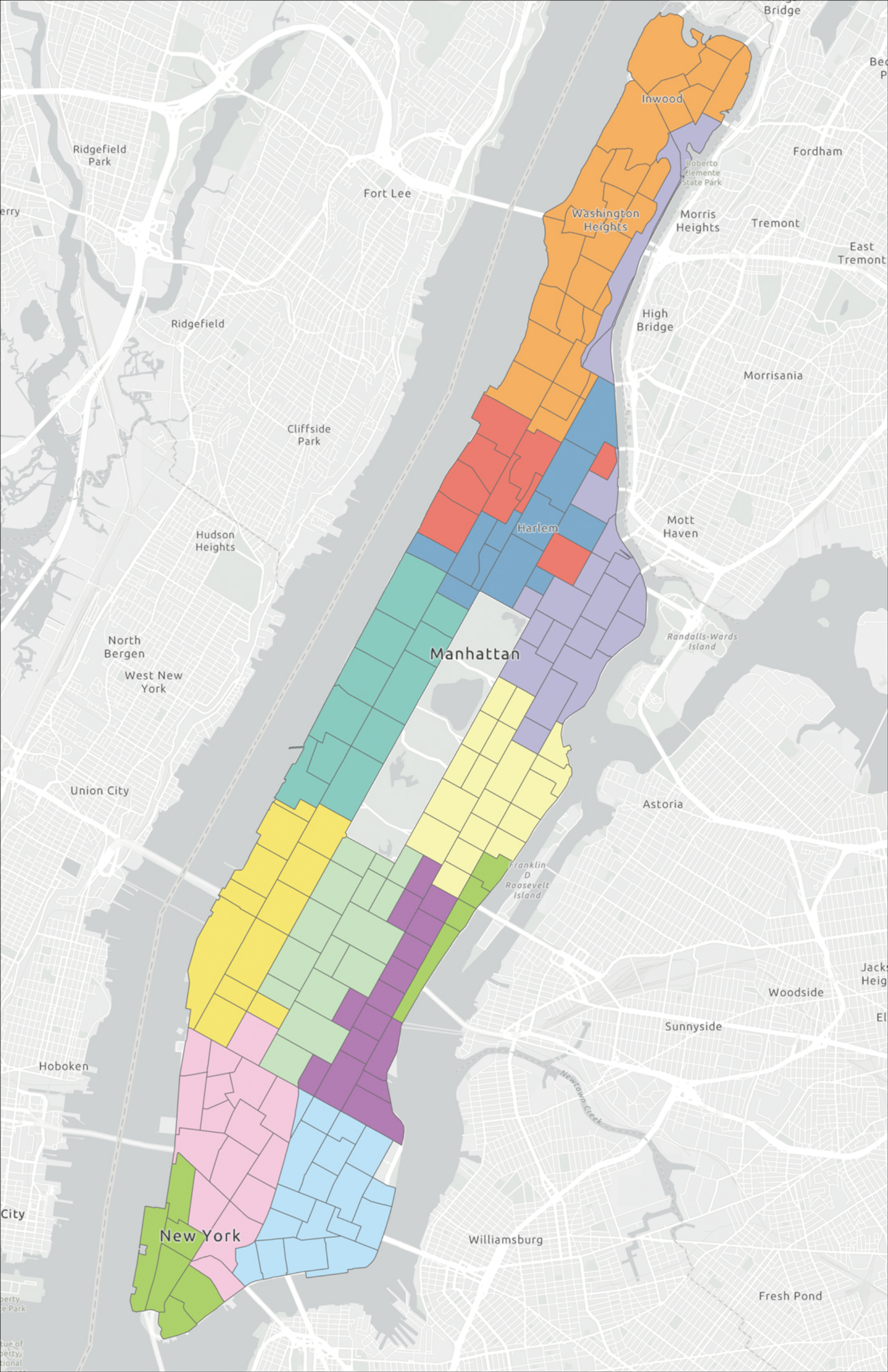}
}
\subfigure[Ours]{
\includegraphics[scale=.11]{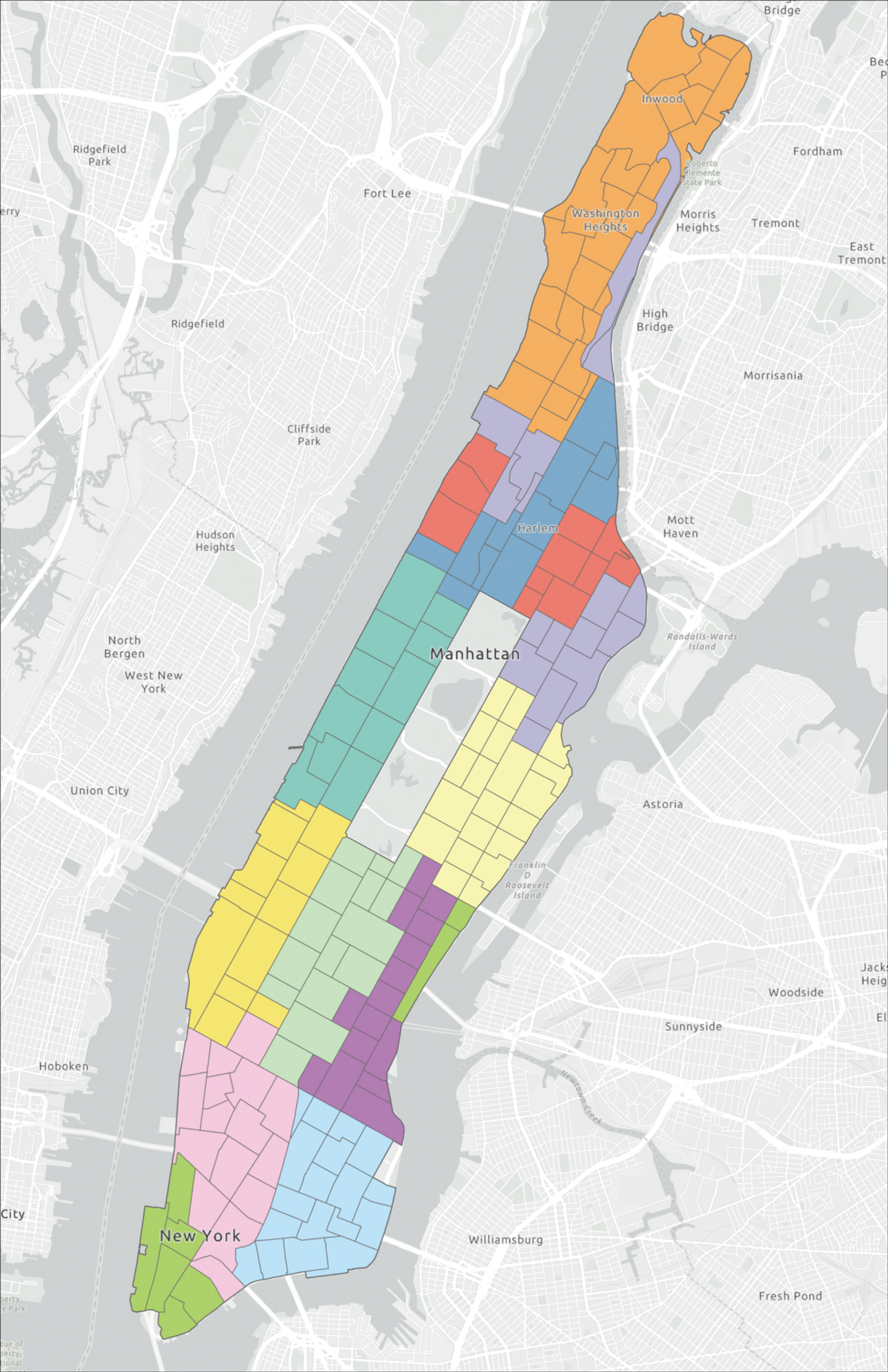}
}
\caption{Districts in Manhattan and region clusters.}

\label{landuse}
\end{figure}

From the observations in Figure.~\ref{abla}, it is evident that the multi-graph aggregation module (w/o MGAM) has the most significant impact on performance. This underscores the crucial role of our improved graph attention mechanism in effectively recognizing dependencies, particularly in long-distance regions for urban region embedding. The second-largest impact comes from the absence of w/o DSGF, affirming its effectiveness in enhancing the learning effect for a single view. Additionally, the impact of the absence of w/o AFM is notable, indicating that our improved linear attention not only enhances region representation learning but also improves learning efficiency. Finally, the significantly poorer performance of w/o GCL compared to ATGRL underscores the effectiveness of soft thresholding applications. These ablation results provide further confirmation of the efficiency of our model in addressing the urban region embedding problem.

\subsubsection{Visualized Analysis}
For the visual assessment of the land usage classification task clustering results, we depict the clustering outcomes of five baselines and our model in Figure.~\ref{landuse}, with regions in the same cluster marked with the same color. Notably, the clustering results obtained by our method show the most ideal degree of consistency with the real boundaries of the ground conditions. This observation suggests that the region embeddings learned by our model more effectively represent regional functions. The superior performance can be attributed to our method's proficiency in capturing long-range relationships among embeddings in distinct urban regions and adeptly integrating information from multi-views.

\begin{table}[!t]
\centering
\caption{Comparison of computation time \& inference time.}
\begin{tabular}{@{}cccc@{}}
\toprule
\multirow{2}{*}{Task}                      & \multirow{2}{*}{Model} & \multicolumn{2}{c}{Computation Time} \\ \cmidrule(l){3-4} 
                                           &                        & Training (s/100epoch)   & inference  \\ \midrule
\multirow{5}{*}{Land usage classification} & MVGRE                  & 33.46                   & 1.207      \\
                                           & w/o GCL                & 31.38                   & 1.232      \\
                                           & w/o MGAM               & 28.83                   & 1.197      \\
                                           & w/o AFM                & 36.35                   & 1.201      \\
                                           & ATGRL                  & 26.83                   & 1.133      \\ \bottomrule
\end{tabular}
\label{time}
\end{table}

\subsubsection{Computational Efficiency Comparison} 
In this section, we compare the training and inference times of MVGRE and the different variants of ATGRL in the land use classification task. It is essential to note that the selection of MVGRE as the baseline model is based on its utilization of the TensorFlow framework and its use of human mobility data, POI data, and check-in data as inputs. This choice allows for better control of variables, facilitating a more accurate comparison of computational time. The comparison results are shown in Table~\ref{time}. Specifically, w/o AFM greatly increases both training and inference time by using the self-attention mechanism instead of the attentive fusion module in ATGRL. Moreover, the outcomes obtained w/o GCL indicate that incorporating soft thresholding induces sparsity in the input features, thereby diminishing computational demands and accelerating the training process.
\begin{figure}
\centering\includegraphics[width=1.0\textwidth]{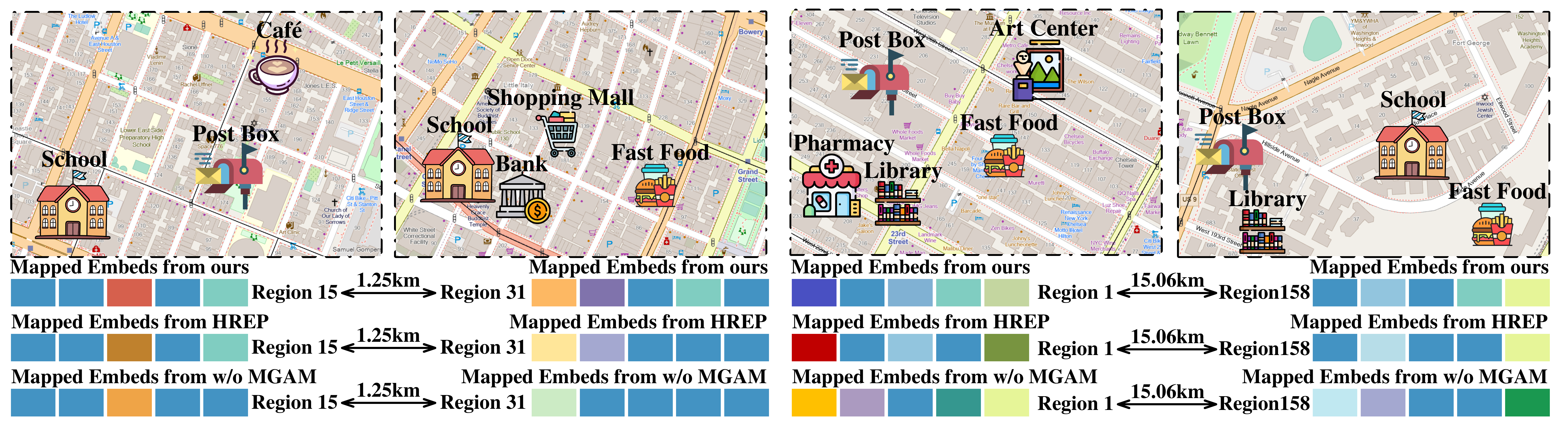}
\caption{Case study of our ATGRL method on New York City datasets.}
\label{case}
\end{figure}
In conclusion, our proposed model, ATGRL, not only improves the effectiveness of downstream tasks but also exhibits superior computational efficiency compared to MVGRE. This improvement can be attributed to the optimized attention mechanism, which replaces the complicated self-attention mechanism used in MVGRE, along with the introduction of an effective noise reduction mechanism.

\subsubsection{Case Study}
In this section, we conduct a case study to demonstrate the capabilities of ATGRL in learning global region dependencies in urban region embedding. Specifically, we sample two pairs of regions: 1) nearby regions, namely (Region-15, Region-31); and 2) more distant regions, denoted as (Region-1, Region-158). As illustrated in Figure. \ref{case}, Region-15 and Region-31 are spatially adjacent, yet they exhibit distinct urban functions. However, the region embeddings acquired through the baseline HERP display a noticeable bias in functional identification. Notably, embeddings obtained using GAT without the multi-graph aggregation module (i.e., w/o MGAM) demonstrate a high degree of similarity. In contrast, embedding diversity can be observed in the region embeddings learned by our ATGRL method. Furthermore, despite the considerable geographical distance between Region-1 and Region-158, our encoding embeddings successfully preserve their fundamental semantics. This is an accomplishment not achieved by HERP and w/o MGAM. In conclusion, these observational results indicate that our ATGRL demonstrates an advantage in capturing global spatial dependencies within urban regions. This is attributed to the operation of the multi-graph fusion module designed in this paper.

\begin{table}[!t]
\centering
\caption{Comparison of the training time and inference time when varying the number of elements \(H\) in a memory unit as defined in Eq.\ref{linearAtt}.}
\begin{tabular}{@{}cccc@{}}
\toprule
\multirow{2}{*}{Task}                      & \multirow{2}{*}{\#H} & \multicolumn{2}{c}{Computation Time} \\ \cmidrule(l){3-4} 
                                           &                      & Training (s/100epoch)   & inference  \\ \midrule
\multirow{6}{*}{Land usage classification} & 8                    & 27.94                   & 1.273      \\
                                           & 32                   & 25.83                   & 1.133      \\
                                           & 64                   & 28.94                   & 1.315      \\
                                           & 128                  & 31.71                   & 1.426      \\
                                           & 256                  & 33.48                   & 1.751      \\
                                           & 512                  & 38.89                   & 1.893      \\ \bottomrule
\end{tabular}
\label{Htime}
\end{table}

\begin{figure}[!t]
\centering
\subfigure[$H$]{
\includegraphics[scale=.25]{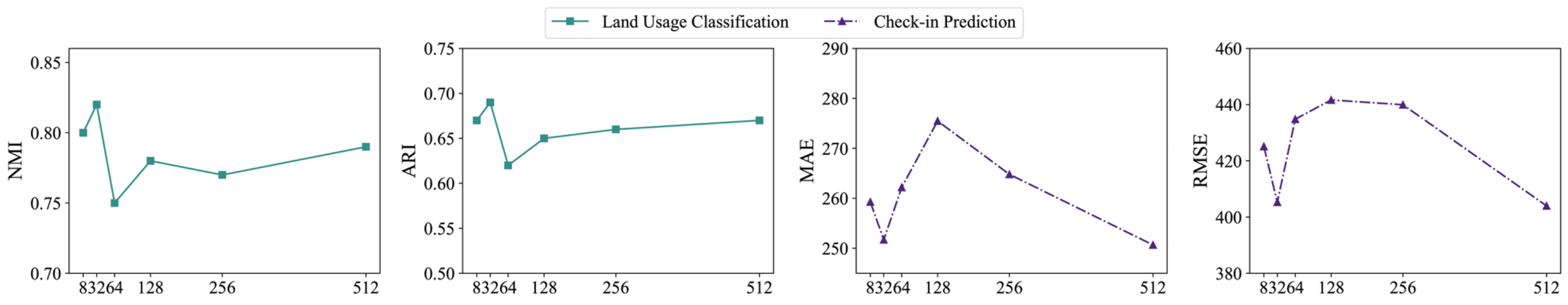}
\label{H}
}
\subfigure[$\beta$]{
\includegraphics[scale=.25]{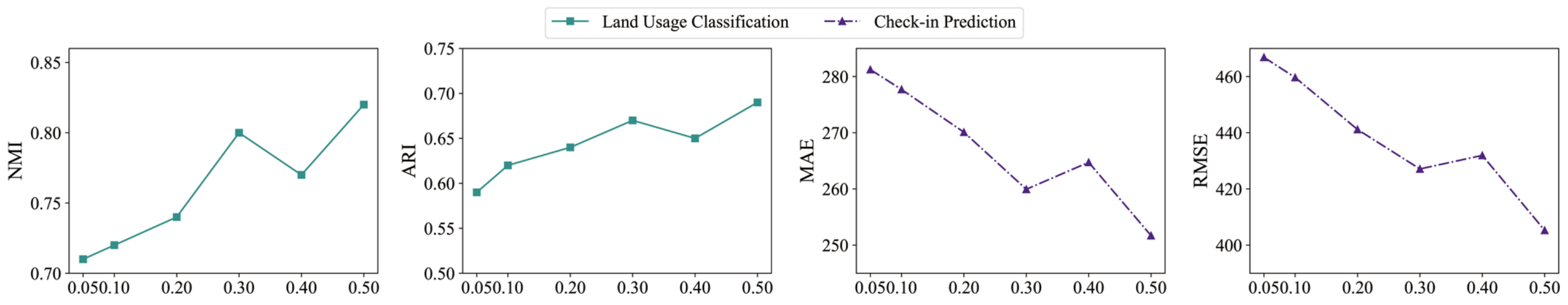}
\label{beta}
}
\caption{Impact of $H$ and $\beta$ to our model.}
\label{hypter}
\end{figure}

\subsubsection{Hyper-parameter Studies}
To show the effect of different parameter settings, we conduct experiments to evaluate the performance of our framework ATGRL with different configurations of important hyper-parameters (e.g.$\# H$ and $\# \beta$). When varying a specific hyper-parameter for effect investigation, other parameters are fixed with default values. The results are shown in Figure.~\ref{hypter}.

First, we analyze how the experimental results vary with different numbers of elements \(H\) in memory units. We tune $H$ in $\{8, 32, 64, 128, 256, 512\}$, and then check the corresponding results. As can be seen in Figure.~\ref{H} and Table.\ref{Htime}, $H$ = 32 is an ideal choice. In fact, \(H\) determines the number of storage units and, consequently, the computational load for calculating the correlation between each view and other views in Eq.\ref{linearAtt}. Increasing \(H\) results in higher computational complexity, allowing the model to scrutinize the relationships between different views in greater detail. A choice of \(H=32\) strikes a balance between computational resources and performance.

Additionally, the hyperparameter $\beta$ affects the result that integrates the global and local region representation in Eq.~\ref{Beta}. We vary $\beta$ from the range of $\{0.05, 0.1, 0.15, 0.2, 0.3,\\ 0.4, 0.5\}$. From Figure.~\ref{beta}, $\beta$ = 0.50 gives the best performance. Balancing global and local region representation is more favorable for urban region embedding in the gated fusion module we designed, which is our best choice.

\section{Conclusion}
\label{sec6}
In this paper, we introduce "ATGRL", a novel approach for efficient and accurate urban region representation. Our method leverages attentive graph-enhanced neural networks to capture both local and global spatial dependencies with heterogeneous and noisy urban data. 
We conduct extensive experiments on three urban downstream tasks with real-world datasets, to validate the effectiveness and versatility of our proposed ATGRL in different settings. In future research, we will apply the proposed framework to other downstream tasks, such as financial and healthcare prediction. 

\bmhead{Acknowledgements}
This work was supported in part by the National Key R$\&$D Program of China under Grant No.2020YFB1710200, in part by the China Postdoctoral Science Foundation under Grant No.2022M711088, in part by the National Natural Science Foundation of China under Grant No.62172243.











\section*{Declarations}


\begin{itemize}
\item Competing interests: All authors declare that they have no competing interests in connection with this manuscript.
\item Authors contribution statement: 
\begin{itemize}
    \item Weiliang Chan: Conceived and designed the study, collected and analyzed data, and drafted the manuscript.
    \item Qianqian Ren: Provided guidance, expertise, and critical revision throughout the research process.
    \item Jinbao Li: Contributed to the research concept, provided valuable feedback, and participated in data interpretation.
\end{itemize} 
All authors approved the final manuscript.
\item Ethical and informed consent for data used: The research presented in this manuscript did not involve any ethical concerns, as it did not include human subjects, animal experiments, or other ethical considerations. Therefore, no formal ethics approval or consent to participate was required.
\item Data availability and access: The datasets generated and/or analyzed during the current study are available at https://opendata.cityofnewyork.
\end{itemize}

\bibliography{sn-bibliography}

\end{document}